%% file: NEC-RP_en.tex
\documentclass[wcp]{jmlr}

\usepackage{longtable}

\usepackage{booktabs}
\usepackage{algorithm}
\usepackage{algorithmic}

\jmlrvolume{101}
\jmlryear{2019}
\jmlrworkshop{ACML 2019}

\title[RP in NEC]{Random Projection in Neural Episodic Control}

 \author{\Name{Daichi Nishio} \Email{dnishio@csl.ec.t.kanazawa-u.ac.jp}\AND
  \Name{Satoshi Yamane} \Email{syamane@is.t.kanazawa-u.ac.jp}\\
  \addr Kanazawa University, Ishikawa, Japan
 }

\editors{Wee Sun Lee and Taiji Suzuki}

\begin{document}

\maketitle

\begin{abstract}
End-to-end deep reinforcement learning has enabled agents to learn with little preprocessing by humans.  
However, it is still difficult to learn stably and efficiently because the learning method usually uses a nonlinear function approximation.
Neural Episodic Control (NEC), which has been proposed in order to improve sample efficiency, is able to learn stably by estimating action values using a non-parametric method.
In this paper, we propose an architecture that incorporates random projection into NEC to train with more stability.
In addition, we verify the effectiveness of our architecture by Atari's five games.   
The main idea is to reduce the number of parameters that have to learn by replacing neural networks with random projection in order to reduce dimensions while keeping the learning end-to-end.

\end{abstract}
 

\input{chapter/chapter1.tex}

\input{chapter/chapter2.tex}

\input{chapter/chapter3.tex}

\input{chapter/chapter4.tex}

\input{chapter/chapter5.tex}

\input{chapter/chapter6.tex}

\input{chapter/chapter7.tex}

\input{chapter/chapter8.tex}

\bibliography{acml19}

\appendix

\input{chapter/appendix.tex}

\end{document}

%% file: chapter/chapter1.tex
\section{Introduction}
Recent advances in deep learning have allowed more complex function approximation and feature extraction.
Reinforcement learning has also benefited from this, and Deep Q-Network (DQN)~\cite{DQN} has been proposed, which enables end-to-end learning from only image features.
However, there are still problems that deep reinforcement learning has to solve.
One of them is poor sample efficiency.
Neural Episodic Control (NEC)~\cite{NEC} has been proposed to solve it.
NEC has introduced a differentiable dictionary, called Differentiable Neural Dictionary (DND), into neural networks.
This makes it possible to learn action values from a feature of a state stably and to learn with a small number of learning steps.
The reason is that it has reduced the parameters which have to learn by changing parametric prediction to non-parametric one.
Furthermore, NEC's architecture allows end-to-end learning, although extraction of features from images is parametric and output action values ​​from them is non-parametric.

In this research, we propose a method that incorporates random projection, which is one of the non-parametric dimensionality reduction methods, into a part of NEC's network for further stable learning.
We also verify its effectiveness in video game experiments.

%% file: chapter/chapter2.tex
\section{Deep Reinforcement Learning}

Reinforcement learning is an area of machine learning that an agent learns how to maximize the return of rewards obtained by interaction with the environment.
The action value of a reinforcement learning agent taking the action \( a \) in the state \( s \) is defined by the following equation (\ref{eq:q-value}), where \(G _ { t } = \sum _ { t } \left( \gamma ^ { t } r _ { t } \right)\) is the sum of discounted rewards, and \( \gamma  \) (\(0 \leq \gamma < 1\)) is the discount factor that represents how important future rewards are.

\begin{equation}
    \label{eq:q-value}
    Q_{\pi} (s, a) = E_{ \pi } \left[ G_{t} | s , a \right]
\end{equation}

The action value is called Q-value, and Q-learning~\cite{q-learning} is a method to estimate the value.
It is a bootstrap estimation using Bellman equation~\cite{Bellman} as in the equation (\ref{eq:bellman-equation}).

\begin{equation}
    \label{eq:bellman-equation}
    Q ( s , a ) \leftarrow Q ( s , a ) + \alpha \left( r + \gamma \max _ { a ^ { \prime } } Q \left( s ^ { \prime } , a ^ { \prime } \right) - Q ( s , a ) \right)
\end{equation}

In value-based policy, the agent chooses an action that maximizes the estimated Q-value.
However, it may not get better rewards for inexperienced states if it continues to choose the greedy action to maximize the value.
In order to avoid this problem, there is a simple but powerful way called \(\varepsilon\)-greedy policy.
It is written as the equation (\ref{eq:greedy}).

\begin{equation}
    \label{eq:greedy}
    \pi ( a | s ) = \left\{ 
        \begin{array} {cl} 
            1 - \varepsilon  & ( a = \operatorname { argmax } _ { a } Q ( s , a )) \\ 
            \varepsilon  & ( \text {otherwise} )
        \end{array} \right.
\end{equation}

Classically, Q-values are recorded in a Q-table, but in reality the state space is large and the action space may be continuous.
Therefore, it is general to approximate an action value function.

Deep learning is also a way of function approximation.
The method which approximates an action value function is generally called deep reinforcement learning.
It is able to extract features of states and to estimate the function which it is difficult for other approximation methods to express.
DQN has attracted much attention for end-to-end learning by extracting embeddings from Convolutional Neural Network (CNN)~\cite{CNN} only based on images and outputting action values in a Fully Connected (FC) layer based on the embeddings.

However, Q-learning by nonlinear function approximation cannot generally guarantee the convergence, and it is difficult to learn stably.
Therefore, DQN uses experience replay~\cite{ER} which creates minibatches by taking experiences randomly, and has a network of the old parameters as the target value network.
In particular, an efficient use of experience has been found to be important for fast and stable learning.
Prioritized Experience Replay (PER)~\cite{PER} has been proposed to weight experiences instead of conventional random sampling.
The paper~\cite{Rainbow} that has examined the importance and combination of various methods has also verified the effectiveness of PER, and it has shown that it has been very important to use experiences efficiently.

%% file: chapter/chapter3.tex
\section{Neural Episodic Control}\label{ch:NEC}

\begin{figure}[htp]
    \begin{center}
        \includegraphics[clip,width=\textwidth]{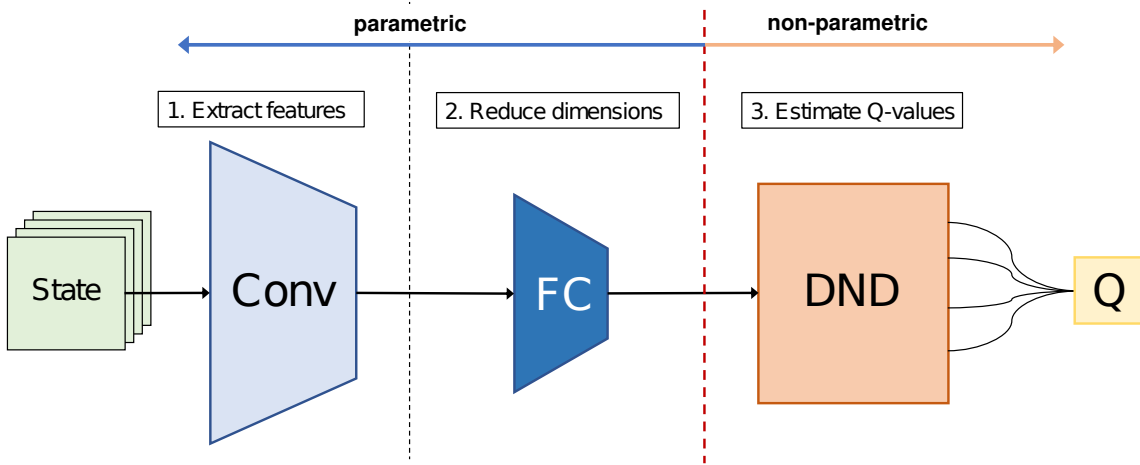}
        \caption{Neural Episodic Control}
    \end{center}
    \label{fig:NEC}
\end{figure}

Neural Episodic Control (NEC)~\cite{NEC} has a memory stored experience inside the architecture to enhance sample efficiency further.
NEC is based on Model-Free Episodic Control (MFEC)~\cite{MFEC}.
The main idea of ​​MFEC is to store many experiences in a Q-table, and it estimates action values by a non-parametric method for embeddings extracted by Variational AutoEncoder (VAE)~\cite{VAE} or random projection.
In contrast, NEC adopts this idea as a part of the architecture by Differentiable Neural Dictionary (DND), which is a dictionary that enables to update by gradients for a Q-table of embeddings.
This makes it possible to learn end-to-end while keeping the Q-table inside the network.

Specifically, the architecture of NEC can be divided into the following three parts.
\begin{enumerate}
    \item \label{NEC1} Obtain an embedding \(\mathbf { h }\) such as by convolutional layers.
    \item \label{NEC2} Obtain an embedding \(\mathbf { h' }\) reduced dimensions of \(\mathbf { h }\) by fully connected layers.
    \item \label{NEC3} Lookup to output action values from \(\mathbf { h' }\) using a k-nearest neighbor algorithm and a kernel function.
\end{enumerate}
Here, we describe~\ref{NEC2} and~\ref{NEC3}.

In~\ref{NEC2}, NEC reduces the embedding dimensions obtained in~\ref{NEC1}.
There are two reasons for doing it, the first is to reduce the space complexity of DND.
The second is the reduction of time complexity of the k-nearest neighbor algorithm in~\ref{NEC3}.

We explain an idea for non-parametric estimation of action values in~\ref{NEC3}.
It is assumed that the \( Q(\mathbf { h' }, a) \) of each action  \( a \) for the embedding \( \mathbf { h' } _ i \) of a state \( s _ i \) similar to the embedding \( \mathbf { h' } \) of a certain state \( s \) should be a similar value in many scenes.
When \(\mathbf { h' }\) obtained in~\ref{NEC2} and the corresponding action \( a \) are input, \( p \)
keys \(\mathbf { h' } _ i\) resembling \(\mathbf { h' }\) among the keys existing in DND are searched by k-nearest neighbor algorithm using kd-trees~\cite{kd}.
Let \( Q _ a \) be the weighted value \( v _ i \) corresponding to the \( p \) keys.

\begin{equation}
    \label{eq:weight}
    w _ { i } = \frac { k \left( \mathbf { h' } ,\mathbf { h'}  _ { i }  \right) } { \sum _ { j } k \left(\mathbf { h' }, \mathbf { h' } _ { j } \right) }
\end{equation}

\begin{equation}
    \label{eq:sum_weight}
    Q _ { a } = \sum ^ {p} _ { i = 1 } w _ { i } v _ { i }
\end{equation}

The function \( k \) in the equation (\ref{eq:weight}) is a kernel function.
For example, it is written as the equation (\ref{eq:inverse}) by the inverse kernel function.

\begin{equation}
    \label{eq:inverse}
    k \left(\mathbf { h' }, \mathbf { h' } _ { i } \right) = \frac { 1 } { \left\| \mathbf { h' } - \mathbf { h' } _ { i } \right\| _ { 2 } ^ { 2 } + \delta }
\end{equation}

Although \( \delta \) is the parameter to prevent division by zero, we should make it little larger such as \(\delta = 10^{-3}\) because each value of \( p \) neighbors is referred.

In this way, NEC allows stable learning by estimating action values non-parametrically.
While it has been possible to learn faster and stabler than many algorithms including DQN, it has been found that long-term training makes it inferior to other methods.
The cause is that it is necessary to store a large number of embedding-action pairs in DND, and the insufficient buffer size cannot estimate action values well.

NEC adopts multi-step Q-learning~\cite{NSQ} as another technique for fast learning.
One-step learning has the advantage that the variance of target value is low, but it also has the disadvantage that the propagation of rewards is slow.
On the other hand, Monte Carlo Q-learning, which uses all the experiences of one episode, has the a rapid propagation of rewards, but unstable learning because of the high variance.
Multi-step Q-learning is responsible for the trade-off the strengths and weaknesses of one-step Q-learning and Monte Carlo Q-learning.

In Rainbow~\cite{Rainbow}, it is better to set the value of this step number \( N \) to 3 or 5, but NEC sets \( N \) to 100 because of the stability of learning.

%% file: chapter/chapter4.tex
\section{Random Projection}\label{ch:RP}

Random Projection (RP) is a linear projection with a random matrix and is used to reduce dimensions of high-dimensional data.
As properties of the projection matrix, it is necessary to consider the time to construct the matrix and the quality of embedding after dimensionality reduction.
The difference between the main methods of random projection is as shown in Table~\ref{tab:RP}.

\begin{table}[htbp]
    \caption{Comparison of random projection methods}
    \begin{center}
    \begin{longtable}{l | c | c | c }
    \hline
    \textbf{RP method} & \textbf{Construction time} & \textbf{Projection time\(^e\)} & \textbf{Embedding quality\(^f\)} \\
    \hline
    Gaussian & \( \mathcal{O} \left(dk\right) \) & \( \mathcal{O} \left( ndk \right) \) & \(\mathcal{O} \left( \varepsilon ^ { - 2 } n \right)\) \\
    Achlioptas' \(^a\) & \( \mathcal{O} \left(dk\right) \) & \( \mathcal{O} \left( ndk \right) \) & no proof \\
    Li's \(^b\) & \( \mathcal{O} \left( \sqrt{d}k \right) \) & \( \mathcal{O} \left( n\sqrt{d}k \right) \) & no proof \\
    SRHT \(^c\) & \( \mathcal{O} \left( dk + d \log d \right) \) & \( \mathcal{O} \left( nd \log k \right) \) & \( \mathcal{O} \left( \varepsilon ^ { - 2 } \left( n + d \right) \log n \right)\) \\
    Count Sketch \(^d\) & \( \mathcal{O} \left( d \right) \) & \( \mathcal{O} \left( nd \right) \) & \( \mathcal{O} \left( \varepsilon ^ { - 2 } n ^ { 2 } \right)\) \\
    \hline
    \end{longtable} \label{tab:RP}
    \(^a\)~\cite{Achlioptas}
    \(^b\)~\cite{Li}
    \(^c\)~\cite{SRHT}
    \(^d\)~\cite{CountSketch} \\
    \(^e\) The projection time is the case of dense inputs. \\
    \(^f\) This quality is oblivious subspace embedding (OSE) lower bound. 
    \end{center}
\end{table}

From here, we describe Gaussian random projection~\cite{GRP} because it has the best \textbf{Embedding quality}.
In this method, a \( m \) dimensional vector \( \mathbf { x } \) is multiplied by the Gaussian random matrix \(\mathbf { R }\) to convert it into a  \(d _ \mathbf { y } \) dimensional vector \(\mathbf { y } \).
\begin{equation}
    \label{eq:RP}
    \mathbf { y }  = \mathbf { R } \mathbf { x } 
\end{equation}

If the elements of the random matrix \( \mathbf { R } \) are generated by random numbers in accordance with a Gaussian distribution (mean 0, variance \(1 / d _ { \mathbf { y } }\)), the distance between the data is approximately maintained with high probability \(1 - \mathcal{O} \left( e ^ { - m \varepsilon ^ { 2 } } \right)\) when any \( N \) training data \(\mathbf { x } ^ { ( j ) } ( j = 1 , \dots , N )\) are projected in the \(d _ { \mathbf { y } = \mathcal{O} \left( \varepsilon ^ { - 2 } \log N \right)}\) dimensions[~\cite{JL-lemma},~\cite{RP-method}].
\begin{equation}
    \label{eq:JL-lemma}
    ( 1 - \varepsilon ) \left\| \mathbf { x } _ { j } - \mathbf { x } _ { i } \right\| ^ 2 _ { 2 } \leq \left\| \mathbf { y } _ { j } - \mathbf { y } _ { i } \right\| ^ 2 _ { 2 } \leq ( 1 + \varepsilon ) \left\| \mathbf { x } _ { j } - \mathbf { x } _ { i } \right\| ^ 2 _ { 2 }
\end{equation}

Gaussian random projection is distinguished by its simplicity and high quality of embedding.
The paper~\cite{expRP} that actually experimented with random projection has shown that the relationship between the dimensions of the vector before projection and the dimensions of the vector after the projection is theoretically guaranteed.
Moreover, it has been applied to the EM algorithm and the performance is improved.
Another paper~\cite{AppRP} applied to an image or text has reported to show good performance even if we has reduced dimensions under weaker conditions than the theoretically guaranteed the inequality (\ref{eq:JL-lemma}).
It has also shown that if the dimension is too small, the accuracy drops sharply.

Random projection has a good property that there is no restriction on the magnitude of each value of data.
This means that the inequality (\ref{eq:JL-lemma}) holds without normalization.

Also, random projection has good compatibility with the kd-trees because information other than Euclidean distance is redundant for the algorithm based on the closeness of the distance.
In other words, we can use it for the data projected by random projection.
In fact, RP-kd-Trees~\cite{RPkd}, which uses random projection for dimension reduction, has been also proposed.

%% file: chapter/chapter5.tex
\section{Related Work}
There are several researches that have improved or combined NEC.
A representative one is Ephemerally Value Adjustments (EVA)~\cite{EVA}.
It has improved the performance by combining NEC and other planning algorithms, and the calculation time for querying DND, which has been a problem with NEC, has also been improved by reducing the number of queries.
Also, there are  combines parametric methods such as DQN with non-parametric methods such as NEC.
Semiparametric Reinforcement Learning (SRL)~\cite{SRL} has proposed a method using action values ​​that combine the values output by neural network and the values ​​estimated by NEC architecture.
NEC2DQN~\cite{NEC2DQN} has also made use of learning efficiency of NEC to assist to learn about a parametric network.
In practical applications, NEC has been applied to vehicular ad hoc networks (VANET), which is an ad hoc network for inter-vehicle communications~\cite{VANEC}, and it has shown improved performance compared to the conventional method.

Deep reinforcement learning using random projection such as MFEC has been proposed.
Episodic Memory Deep Q-Netwroks (EMDQN)~\cite{EMDQN} has regularized action value ​with the values ​​output using random projection.
However, they differ from our proposed algorithm in that they use random projection outside their neural networks, in other words, they are not end-to-end networks.

More recent research has proposed an architecture using random projection for deep neural networks~\cite{RPDL}.
Although the networks train on the embeddings projected by random projection, we do not use it as the input of neural networks, but utilize the relationship of the embedding distance.

%% file: chapter/chapter6.tex
\section{Proposed Algorithm}\label{ch:PA}
As NEC has shown, non-parametric methods such as the k-nearest neighbor algorithm are good for increasing the learning speed of agents because there are no parameters which have to learn.
NEC reduces the dimensions of embeddings output by CNN with FC layers, calculates Euclidean distance of nearby embeddings using k-nearest neighbor algorithm, then outputs the action value.
However, what is important for the inverse kernel function to calculate the value ​​is that the relationship of the distance is maintained as in the equation (\ref{eq:inverse}).
Therefore, we perform stable learning by incorporating random projection which is a non-parametric method and can keep the Euclidean distance relationship.
At the same time, we aim to make the proposed architecture more general by end-to-end learning.

We propose NEC-RP, which incorporates random projection into the architecture of NEC.
We show the algorithm in Algorithm~\ref{alg:NEC-RP}.

\begin{algorithm}[htp]
    \caption{NEC-RP}\label{alg:NEC-RP}
    \begin{algorithmic}[1]
    \STATE Initialize a replay memory \(D\) to capacity \(C_D\).
    \STATE Initialize DND memories \(M_a\) to capacity \(C_{M_a}\).
    \STATE Initialize an action value function \(Q\) with random weights.
    \STATE Initialize the number of entire timesteps \(TS\) to zero.
    \STATE Define the step \(CS\) to replace an RP layer with a FC layer.
    \COMMENT{if \(CS \gets \infty \), use only the RP layer forever.}
    \FOR{each episode}
        \FOR{\(t = 1,2,\ldots,T\)}
            \STATE Receive an observation \(s_t\) from an environment.  
            \STATE Convert \(s _ t\) to an embedding \(h _ t\).
            \IF{\(TS < CS\)} 
                \STATE Reduce dimensions of \(h _ t\) using an RP layer.
            \ELSE 
                \STATE Reduce dimensions of \(h _ t\) using a FC layer.
            \ENDIF
            \STATE Calculate \(Q(s _ t, a)\) from  \(M _ a\).
            \STATE \(a _ t \gets \epsilon\)-greedy policy based on \(Q(s _ t, a)\).
            \STATE Take an action \(a _ t\), receive a reward \(r _ t\).
            \STATE Train on a random minibatch from \(D\).
            \STATE \(TS \gets TS + 1\)
        \ENDFOR
        \FOR{\(t = 1,2,\ldots,T\)}
            \STATE Append \((s_t, a_t, Q^{(N)}(s_t, a_t))\) to \(D\).
            \STATE Append \((h_t, Q^{(N)}(s_t, a_t))\) to \(M _ { a _ t }\).
        \ENDFOR
    \ENDFOR
    \end{algorithmic}
\end{algorithm}

\subsection{Random Projection layers}
Here we introduce Random Projection (RP) layer.
Random projection is a linear projection and differentiable as in the equation (\ref{eq:RP}).
The partial derivative of it is given by the equation (\ref{eq:partialRP}).

\begin{equation}
    \label{eq:partialRP}
    \frac{\partial \mathbf { y } }{\partial \mathbf { x }}  = \mathbf { R ^ T }
\end{equation}

In addition, we can regard the equation (\ref{eq:RP}) as in the following neural network's equation (\ref{eq:NN}) by substituting the random matrix \(\mathbf { R }\) for \(\mathbf { W }\) and a zero vector for \(\mathbf { b }\).

\begin{equation}
    \label{eq:NN}
    \mathbf { y }  = \mathbf { W } \mathbf { x } + \mathbf { b }
\end{equation}

In other words, we define our RP layer by Gaussian random projection as a layer of which the weight matrix is ​​generated by Gaussian random (mean \( 0 \), variance \(1 / d _ { \mathbf { y } }\)) and the bias is a zero vector.
Therefore, we regard our RP layer as a special case of the FC layer.
We show the difference between these layers in Table~\ref{tab:FCandRP}.

\begin{table}[htbp]
\caption{The differences between a FC layer and a RP layer}
\begin{center}
\begin{tabular}{c|c|c}
\hline
\textbf{ } & \textbf{FC} &\textbf{RP} \\
\hline
Initializer (weight) & \begin{tabular}{c}
e.g. Gaussian \\ (Mean=0, \\ Variance=1)
\end{tabular} & \begin{tabular}{c}
e.g. Gaussian \\ (Mean=0, \\ Variance=\(|units_{out}| ^ {-1}\)) 
\end{tabular}\\
\hline
Initializer (bias) & e.g. zeros & zeros \\
\hline
Parameters & update & fix \\
\hline
\end{tabular}
\label{tab:FCandRP}
\end{center}
\end{table}

\subsection{Neural Episodic Control with a Random Projection layer}

\begin{figure}[htp]
    \centering
    \includegraphics[clip,width=\textwidth]{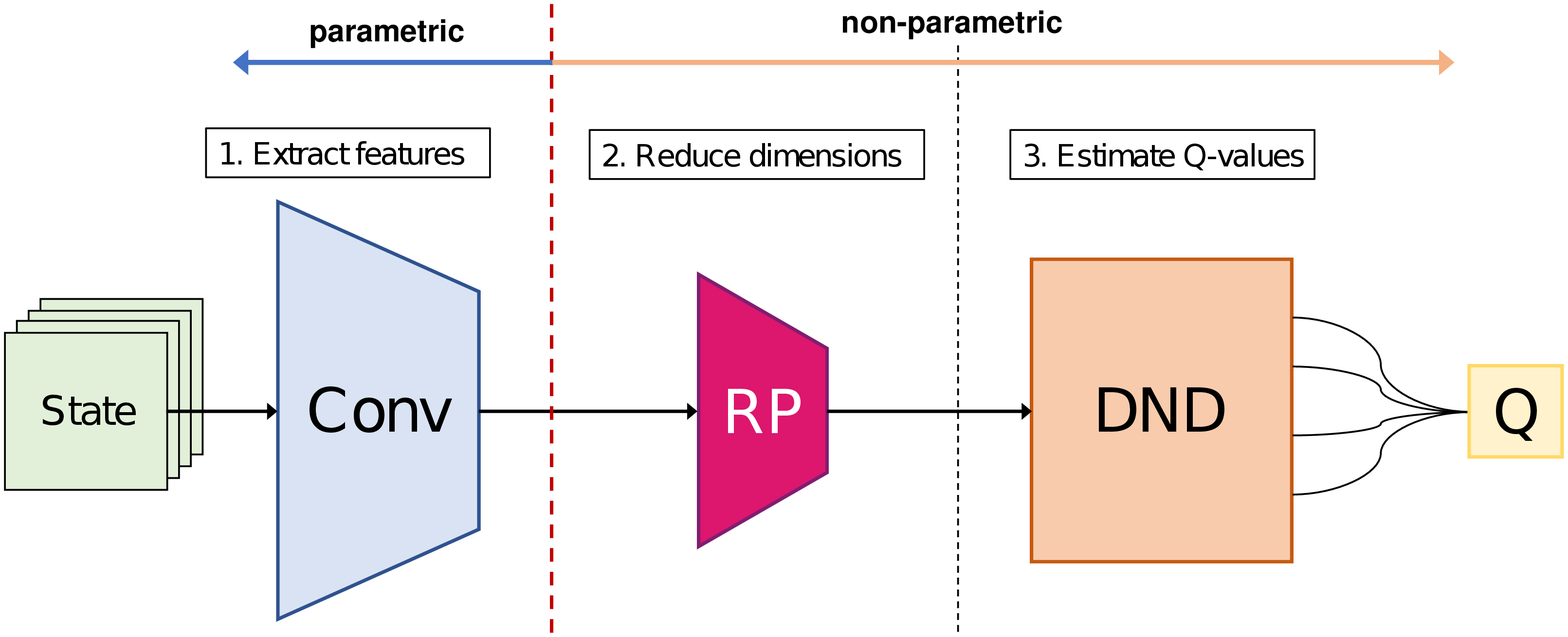}
    \caption{A random projection layer in Neural Episodic Control architecture}
    \label{fig:NEC-RP}
\end{figure}

We show NEC-RP architecture in Fig.~\ref{fig:NEC-RP}.
Of the three parts of the NEC architecture mentioned in chapter~\ref{ch:NEC}, we use the RP layer in~\ref{NEC2}.
In this way, the value of inverse kernel function is approximately maintained as in the inequality (\ref{eq:inverseRP}) from the inequality (\ref{eq:JL-lemma}) with \( \delta \) which is a constant that prevents division by zero.

\begin{eqnarray}
    \label{eq:inverseRP}
    ( 1 - \varepsilon ) \left\| \mathbf { h } - \mathbf { h } _ { i } \right\|  ^ { 2 } _ { 2 } \leq& \left\| \mathbf { h' }  - \mathbf { h' } _ { i } \right\| ^ { 2 } _ { 2 } &\leq ( 1 + \varepsilon ) \left\| \mathbf { h }  - \mathbf { h } _ { i } \right\|  ^ { 2 } _ { 2 } \nonumber \\
    ( 1 - \varepsilon ) \left\| \mathbf { h }  - \mathbf { h } _ { i } \right\| ^ { 2 } _ { 2 } + \delta \leq& \left\| \mathbf { h' }  - \mathbf { h' } _ { i } \right\| ^ { 2 } _ { 2 } + \delta &\leq ( 1 + \varepsilon ) \left\| \mathbf { h }  - \mathbf { h } _ { i } \right\| ^ { 2 } _ { 2 } + \delta \nonumber \\
    \frac { 1 }{( 1 + \varepsilon ) \left\| \mathbf { h }  - \mathbf { h } _ { i } \right\| ^ { 2 } _ { 2 } + \delta} \leq& \frac { 1 }{\left\| \mathbf { h' }  - \mathbf { h' } _ { i } \right\| ^ { 2 } _ { 2 } + \delta} &\leq \frac { 1 }{( 1 - \varepsilon ) \left\| \mathbf { h }  - \mathbf { h } _ { i } \right\| ^ { 2 } _ { 2 } + \delta} \nonumber \\
    \frac { 1 }{( 1 + \varepsilon )\left\| \mathbf { h }  - \mathbf { h } _ { i } \right\| ^ { 2 } _ { 2 } + \delta} \leq& k (\mathbf { h' },\mathbf { h'}_i) &\leq \frac { 1 }{( 1 - \varepsilon ) \left\| \mathbf { h }  - \mathbf { h } _ { i } \right\| ^ { 2 } _ { 2 } + \delta} 
\end{eqnarray}

The value of the kernel function may be unstable if we use parametric and nonlinear function approximation methods such as deep neural network, but in NEC-RP it is guaranteed as in the inequality (\ref{eq:inverseRP}), hence it is possible to learn stably.
However, we should not reduce the dimensions too much because the expressiveness may be insufficient alike neural networks, or the range of the value guaranteed by the inequality (\ref{eq:inverseRP}) may become too wide.

Moreover, our architecture is end-to-end because the entire network is differentiable.
Therefore, it is available to use like NEC.

\subsection{Replacing a Random Projection layer with a Fully Connected layer}

The RP layer does not learn its parameters as shown in Table~\ref{tab:FCandRP}.
Although non-parametric methods have the advantage of not requiring learning, they have the disadvantage that the performance is inferior to parametric ones as the learning progresses.
However, if we modify the parameters of the RP layer to learn, we can consider it as a FC layer because the RP layer is possible to learn from the middle of training.
For example, the simplest method is to switch the RP layer to a FC layer according to the time step \( t \) and the hyperparameter \( CS \).

\begin{equation}
    \label{eq:RP2FC}
    \mathbf { h' } = \left\{ 
        \begin{array} {cl} 
            f_{RP} ( \mathbf { h } ) & ( t < CS ) \\ 
            f_{FC} ( \mathbf { h } ) & ( \text {otherwise} )
        \end{array} \right.
\end{equation}

\begin{figure}[htp]
    \centering
    \includegraphics[clip,width=\textwidth]{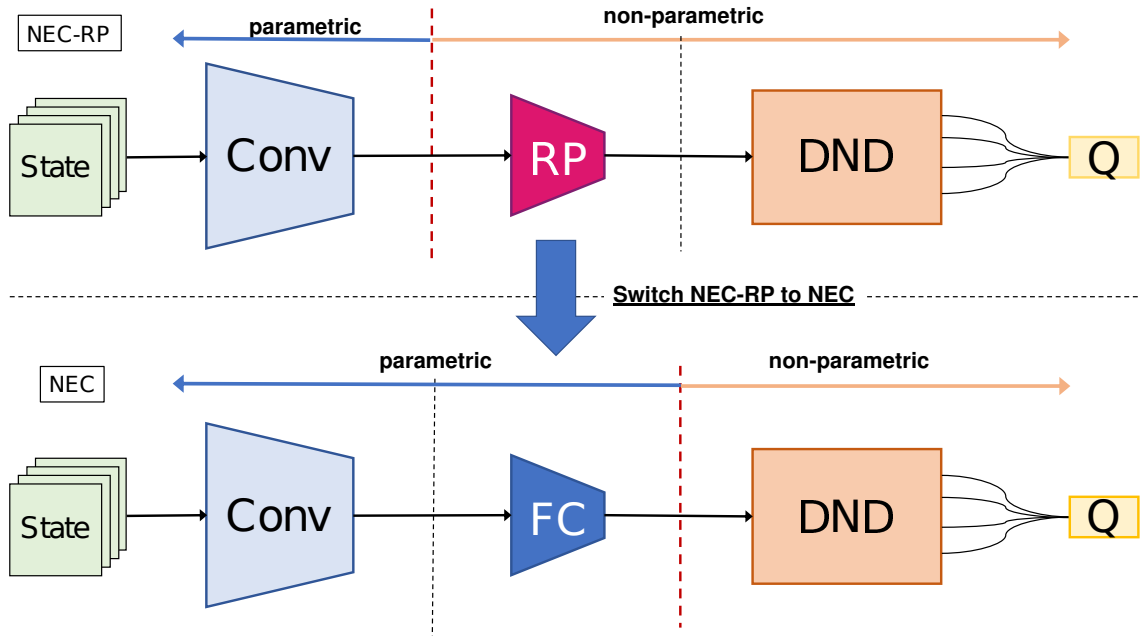}
    \caption{Changing an RP layer of NEC-RP to a FC layer}
    \label{fig:RP2FC}
\end{figure}

This switching is the same as pre-training the other parameters than the layer for dimensionality reduction (Fig.~\ref{fig:RP2FC}).

\subsection{Initialization}

The initialization of the convolution layer is the same as DQN and NEC.
As for a random matrix in the RP layer, we should consider from the Table~\ref{tab:RP}, but in NEC-RP we consider that \textbf{Embedding quality} is the most important factor.
Therefore, we adopt Gaussian random projection.
In other words, we use a random matrix \( \mathbf { R } \) generated by Gaussian random with mean \( 0 \), variance \(1 / d _ { \mathbf { y } } \), where \( \mathbf { y } \) is the vector output by the previous layer.
Naturally the bias is a zero vector.

%% file: chapter/chapter7.tex
\section{Experiments}\label{ch:Exp}

Alike DQN and NEC, we experiment with Atari2600, a video game environment provided by OpenAI Gym~\cite{OpenAI}.
As for baseline NEC, we use the one reproduced in reinforcement learning library Coach~\cite{coach}, and implement NEC-RP~\footnote{Our implementation is available on \href{https://github.com/dnishio/NEC-RP}{https://github.com/dnishio/NEC-RP}.} with this framework.
We also use Faiss~\cite{faiss} as approximate nearest neighbor library.
The target video games are \{\textit{MsPacman, SpaceInvaders, Bowling, Boxing, DoubleDunk}\}.
We experiment three times with different random seeds for each game, on the other hand, we use a fixed seed for random matrix values ​​in our RP layer.
In each experiment, our agent learns for 10M frames.
In order to prevent the learning from becoming unstable due to the variation in the scale of the rewards of the games, DQN clips the rewards \( r \) to \( -1 \leq r \leq 1 \).
However, clipping them affects the final sum of total rewards because high rewards are treated the same as small rewards [\cite{pop-art1},~\cite{pop-art2}].
NEC-RP also does not clip them because NEC achieved stable learning without it.
The parameters such as CNN and DND are the same as NEC.
We show the details of our experiment parameters and our architecture in Table~\ref{tab:exp},~\ref{tab:net} in Appendix~\ref{ap:experiment}.

\subsection{NEC vs. NEC-RP}
\begin{figure}[htp]
    \centering
    \includegraphics[clip,width=0.9\textwidth,height=0.5\textheight]{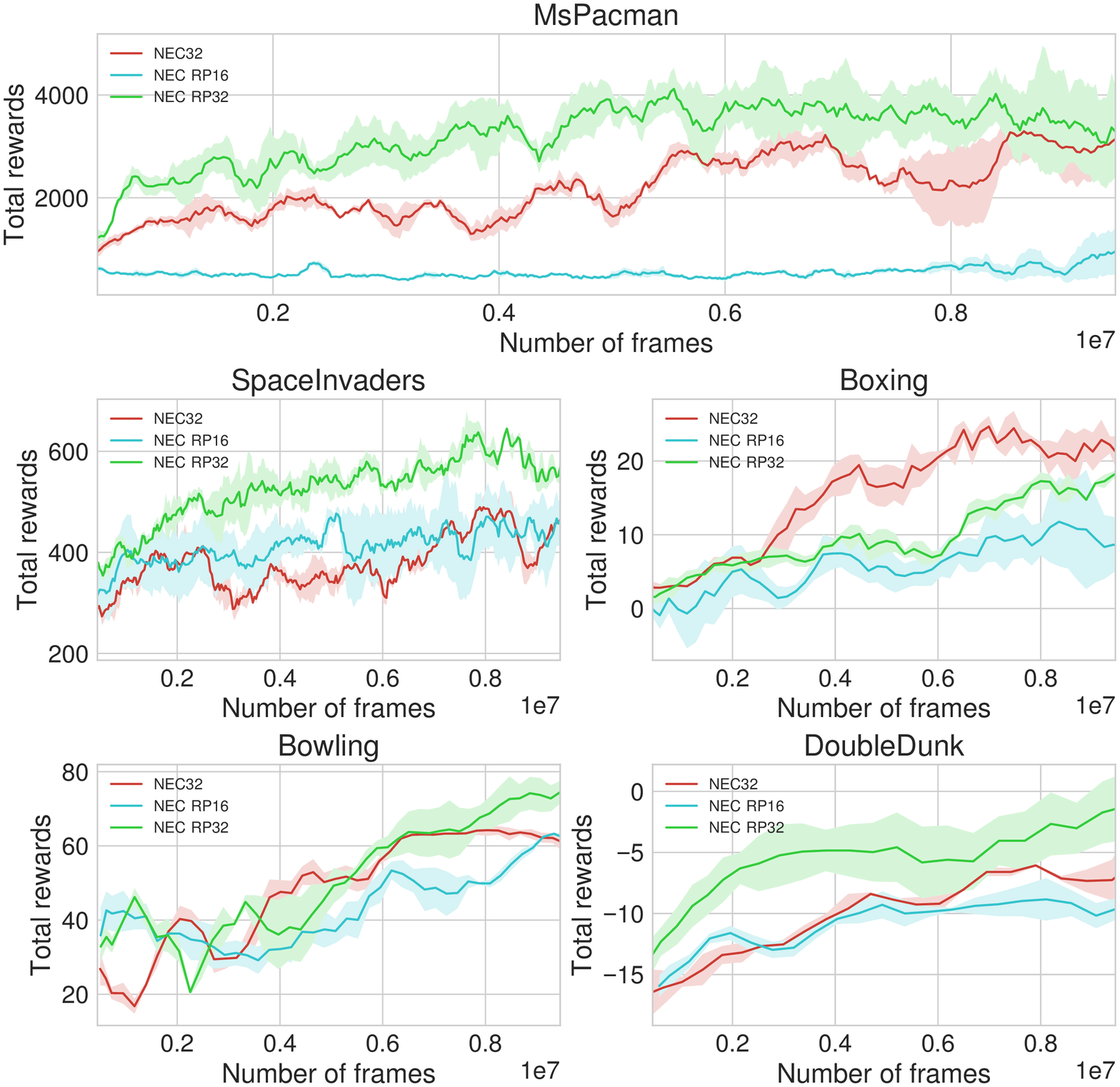}
    \caption{The result of NEC vs. NEC-RP}
    \label{fig:NECvsNEC-RP}
\end{figure}

Here we compare NEC and NEC-RP.
NEC reduces the embedding dimensions to 32 dimensions (NEC32), and NEC-RP reduces to 32 dimensions (NEC-RP32) and reduces to 16 dimensions (NEC-RP16).
In the comparison of NEC32 and NEC-RP32, we verify whether our proposed architecture that introduced random projection outperforms the performances of NEC.
Additionally, in the comparison between NEC-RP32 and NEC-RP16, we examine how the wide range of the inequality (\ref{eq:inverseRP}) affects their performances.

We have shown the result in Fig.~\ref{fig:NECvsNEC-RP}.
Comparing NEC32 and NEC-RP32, we have found that NEC-RP has obtained higher scores in the case of four out of five games.
From this fact, we have considered that not only it has been possible to introduce random projection into NEC, but also NEC-RP has learned faster because the parameters required to update have been reduced.
In particular, we have shown that it has been possible to stably learn even in the games with the rewards \(r > 1\) such as \textit{MsPacman} and \textit{Bowling}.
However, in \textit{Boxing} performance of NEC has been superior to ours in the early stages of learning.
This has implied that there have been some tasks have been difficult for NEC-RP.

We have also found that random projection has had bad performance if we reduce embedding dimensions too much.
In fact, we have seen that the score has been worse in all five games when we have reduced the dimensions to 16 dimensions.
Since NEC-RP uses k-nearest neighbor algorithm, the computation time can be reduced when we reduce the dimensions as small as possible, but the performance is sacrificed.
Therefore, it is important to consider the tradeoff.

\subsection{Switching NEC-RP to NEC}
\begin{figure}[htp]
    \centering
    \includegraphics[clip,width=0.9\textwidth,height=0.5\textheight]{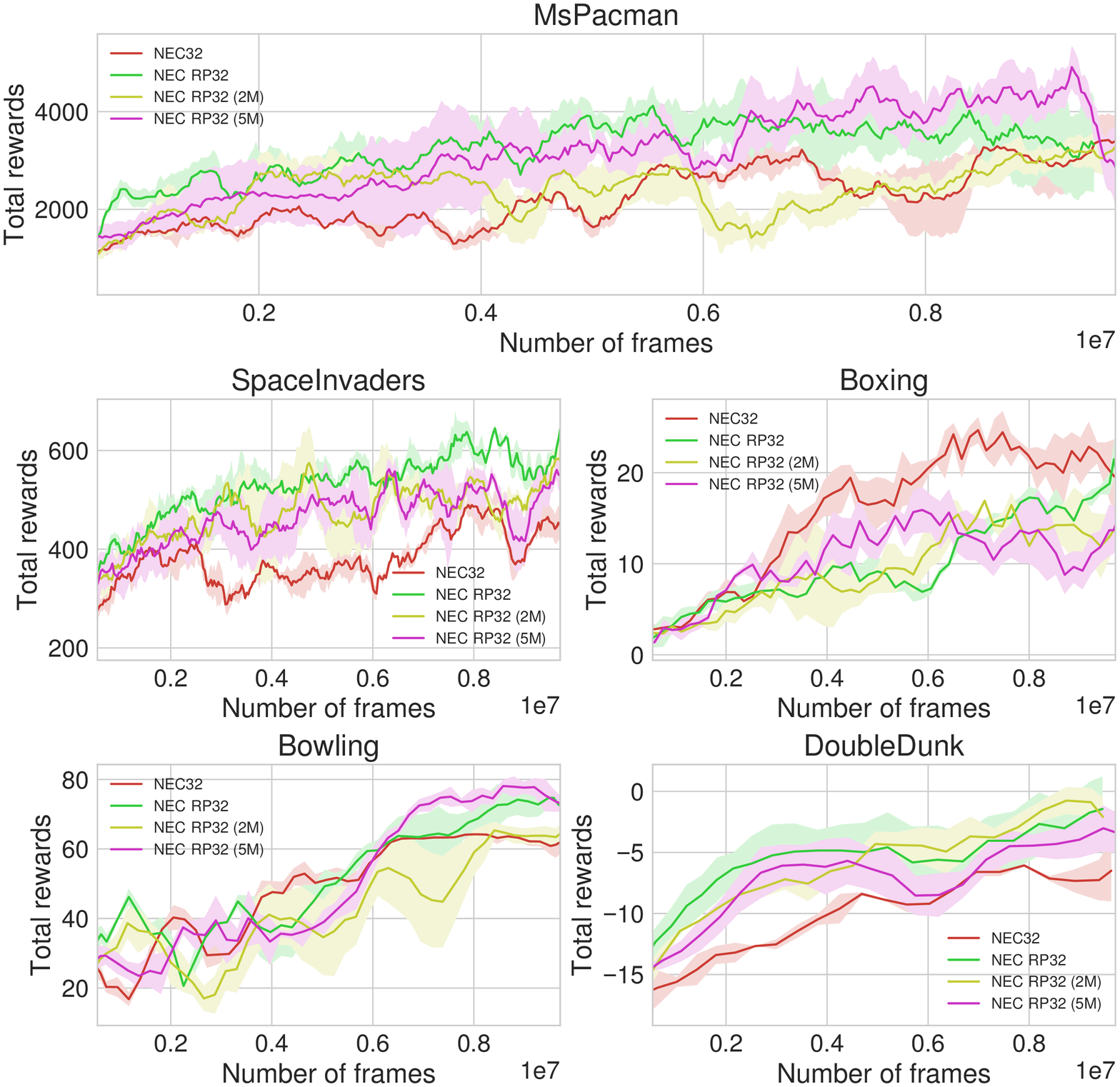}
    \caption{The result of switching NEC-RP to NEC}
    \label{fig:RP2NEC}
\end{figure}

We compare the performance by switching NEC-RP to NEC, that is, we switch the RP layer to a FC layer with a heuristic step count \( CS \).
This comparison indicates experimentally that it is possible to switch from NEC-RP to NEC, and we also verify whether the flexibility of learning about the neural networks can reduce the losses that random projection cannot minimize.
We fix the embedding reduced dimensions to 32 dimensions, and we experiment in the case of switching in 2M frames (NEC-RP32 (2M)) and in 5M frames (NEC-RP32 (5M)).

We have shown the result in Fig.~\ref{fig:RP2NEC}. 
Even when we have switched it in 2M frames or in 5M frames, it has learned without a sharp drop in their performances.
From the result, we have found that it has been possible to switch the RP layer to a FC layer at any time.
However, the performance has been about the same as NEC such as \textit{MsPacman} and \textit{Bowling} at 10M frames if the timing of switching is too early.

On the other hand, in the case of switching in 5M frames, we have shown that the result of exceeding the NEC-RP's performance in \textit{Bowling} even though it has been the architecture similar to NEC at 10M frames.
We have also shown similar performance to NEC-RP in the other games, and we have found that it has been effective to use the RP layer only at the beginning of NEC's training.
However, in \textit{MsPacman} we have observed that the score has dropped sharply around 10M frames.
It is necessary to confirm whether we see this phenomenon in other games in long-term experiments in the future.

%% file: chapter/chapter8.tex
\section{Conclusion}

In this research, we have proposed NEC-RP as the more stable and efficient learning architecture than NEC.
We have experimented with the Atari games and actually have outperformed the performance in four out of five games and have shown that our agent has learned efficiently.
In addition, we have experimented to switch from NEC-RP to NEC, and we have found that it is possible to improve the performance by switching an RP layer to a FC layer.

As future work, we should experiment with long-term training.
In NEC's paper, there is a report that NEC the performance is inferior to DQN one as learning progresses.
We consider the reasons are the limited space of DND and the end of non-parametric methods.
We also need to verify performance of NEC-RP by long-term training because it may be inferior to DQN like NEC.

Moreover, our architecture is easily available for other architectures that use NEC, and we can expect to improve the performance.
We would like to research whether we can apply it to these methods in the future.

%% file: chapter/appendix.tex
\section{Experimental details}\label{ap:experiment}

\begin{table}[htbp]
\caption{Hyperparameters}
\label{tab:exp}
\begin{center}
\begin{tabular}{l | c}
\hline
\textbf{Parameter}&\textbf{Value} \\
\hline
Optimizer & Adam \\
Optimizer learning rate & 0.00001 \\
$\varepsilon$ for exploration & 1 \( \to \) 0.01 over 200K frames \\
Replay buffer size & 100,000 \\
DND learning rate & 0.1 \\
DND size & 500,000 per action \\
\( p \) for KDTree & 50 \\
\( N \) for multi step returns & 100 \\
Image shapes & \(84 \times 84 \times 4 \) \\
Replay period & every 4 training steps \\
Minibatch size & 32 \\
Discount rate & 0.99 \\ 
Timestep \( TS \) for switching NEC-RP to NEC &  2M or 5M frames \\
Entire training frames & 10M frames \\
Heatup &  50,000 frames \\
Evaluation interval &  every 100 episodes \\
$\varepsilon$ for each evaluation &  0.01 \\
\hline
\end{tabular}
\end{center}
\end{table}

\begin{table}[htbp]
\caption{Network parameters}
\label{tab:net}
\begin{center}
\begin{tabular}{l | c}
\hline
\textbf{Parameter}&\textbf{Value} \\
\hline
CNN channels & 32, 64, 64 \\
CNN filter shapes & \(8 \times 8, 4 \times 4, 3 \times 3 \)\\
CNN strides & 4, 2, 1 \\
Embedding size of NEC & 32 \\
Embedding size of NEC-RP & 32 or 16 \\
Random seed for RP layer & 240 \\ 
\hline
\end{tabular}
\end{center}
\end{table}